\documentclass{jdmdh}
\usepackage{array}
\usepackage{pgfplots}
\pgfplotsset{compat=1.18}

\usepackage{CJKutf8}
\usepackage[english]{babel}
\usepackage[utf8]{inputenc}
\usepackage{graphicx,subfigure}
\titlespacing*{\section}
{0pt}{2.5ex plus 1ex minus .2ex}{1.3ex plus .2ex}
\titlespacing*{\subsection}
{0pt}{1.0ex plus 1ex minus .2ex}{1.3ex plus .2ex}
\titlespacing*{\subsubsection}
{0pt}{1.0ex plus 1ex minus .2ex}{1.3ex plus .2ex}

\title{The Effects of Political Martyrdom on Election Results: \\The Assassination of Abe.}
\author[1]{Miu N. Takagi}
\affil[1]{Waseda University, Japan} 

\corrauthor{Miu N. Takagi}{miu.n.takagi@toki.waseda.jp}

\begin{document}

\maketitle

\abstract{In developed nations assassinations are rare and thus the impact of such acts on the electoral and political landscape is understudied. In this paper, we focus on Twitter data to examine the effects of Japan's former Primer Minister Abe's assassination on the Japanese House of Councillors elections in 2022. We utilize sentiment analysis and emotion detection together with topic modeling on over 2 million tweets and compare them against tweets during previous election cycles. Our findings indicate that Twitter sentiments were negatively impacted by the event in the short term and that social media attention span has shortened. We also discuss how "necropolitics" affected the outcome of the elections in favor of the deceased's party meaning that there seems to have been an effect of Abe's death on the election outcome though the findings warrant further investigation for conclusive results..}


\section{Introduction}


This paper focuses on the Japanese election conducted on 2022, July 10 and the effects the assassination of former Primer Minister Shinzo Abe had on both the political and the electoral landscape. The aim is to determine to use this exceptional case study to observe whether Abe’s passing had any influence on voter sentiment during the 2022 Japanese House of Councillors election, held to elect 125 of 248 members of the upper house of the National Diet of Japan. 

Twitter has gained much attention over the past few years as a source to gain insight into the sentiments of people during election seasons, including research by \citep{hasan2018machine, wang2012system, bermingham2011using}, where they studied effective methods to gain insight into public sentiment through Twitter. 
Most notably, in terms of political opinion mining, \citet{Tumasjan} published a paper on how the number of party mentions accurately reflected the election results for the 2009 federal election in the US. Meanwhile, \citet{Wang} have developed a system to analyze online Twitter sentiments towards electoral candidates for the 2012 US election to provide the public with another way of viewing the political landscape live. The methodologies for studying online microblogging sentiments have developed rapidly, where, for example, \citet{Hasan} have suggested a machine learning and lexicon-based hybrid approach in classifying sentiments while comparing  different machine learning techniques. Others have used purely data-driven approaches \citep{severyn2015twitter,tan2022roberta}, purely lexicon-based approaches \citep{teodorescu2022frustratingly,bose2019analyzing}, or for example hashtags to create labels for supervised learning \citep{abdul2017emonet} and many other methods \citep{giachanou2016like,zhang2011combining}.

While individual analyses have been conducted on various countries, US elections dominate the research conducted on this topic. The analyses on the effects of sudden events, such as the assassination of former cabinet members and the like on elections have not been attempted using quantitative approaches to the best of our knowledge. As such, this paper will first explore the sentiment analysis conducted for countries across the globe as a review of general findings and trends found thus far. It will then compare them to Japan's results as a whole.

We use sentiment analysis and emotion detection to compare the changes in emotional polarity for a period of 18 days before Japan's House of Councillors Election dates since 2006. Following this, topic modeling was conducted to detect specific topics that arose in conjunction with emotional change following former Japanese Prime Minister Shinzo Abe's assassination.

Key hypotheses that guided this research were the following two statements. 

{\setlength{\parindent}{0pt} 
\textbf{\underline{Key Hypotheses}} 
}

\begin{enumerate}
    \item Shinzo Abe's assassination affected the sentiments of tweets regarding elections until election day
    \item Abe's death during and following elections has set a precedent for necropolitics in other nations
\end{enumerate}

\section{Related Works}
\subsection{Sentiment Analysis of Election-related Social Media}

Following \citet{Jungherr}’s systematic review of research methods to examine Twitter use in election campaigns, which identified two main methods of data collection - scraping the platforms by oneself with scripts or using 3rd party software - there has been more research conducted on elections and social media.

Twitter research during elections has been justified by researchers such as \citet{ceron}; they illustrated social media analysis can improve the predictive ability to forecast electoral results. What's more, they have also found that there has been some correlation between social media and the results of traditional mass surveys in Italy and France. 

\citet{Stier} conducted research on political communication, with a focus placed on politicians' use of Twitter. With their notable finding on sparse policy talk, this may also be noteworthy to consider in relation to the Japanese 2022 elections. Social media and mass media were found to have \textit{agenda-setting} effects on each other, aligning with the results of \citet{neuman}. In other words, news taken up by mass media was found to affect social media and vice versa. As will be later discussed in this paper, topics discussed by Japanese Twitter users in earlier election periods seemingly discussed policies while those that came later discussed more about specific candidates. While this seems to be at odds with the results of \citet{Stier}, their work was conducted during the German federal election campaign in 2013. This suggests that the Japanese electoral landscape might be different from those observed in Germany or that there have been new developments in how people tweet about elections. Furthermore, their study targeted candidates and their direct audiences, while this study targets all those who have tweeted about the election. As such, this difference in sampling target may also explain the difference in results observed.

On the Japanese election side, research has been conducted on predicting election results for the Japanese House of Councillors election in 2019 based upon the sentiment analysis of Twitter replies to Candidates \citep{okimoto}. While their results showed that sentiments ranked in the bottom half in terms of importance in predicting elections among various variables such as friends, followers, and verifications, this does not mean that sentiments are not important variables that reflect the ongoing election. As stated by the researchers; the analysis of Twitter sentiments may be effective in predicting election results. 

Aside from research conducted on Twitter, other microblogging and social media sites, such as Facebook and Reddit, have also been scraped for various research projects that use sentiment analysis as a part of their methodology \citep{neri2012sentiment, yu2013impact, kouloumpis2011twitter}. 

Reddit has become a popular platform to discuss politics among American Millennials \citep{roozenbeek}. Capitalizing upon this, unlike other researchers who conduct sentiment analysis on content published onto the internet by users, \citet{roozenbeek} instead examined the sentiments of news to view how subscription to sub-forums on Reddit corresponded to the release of those news during the 2016 US presidential elections. They found that negative news negatively affected the subscription trends for Bernie Sanders and Hillary Clinton, while Donald Trump's subreddit was not affected at all. While conducting research similar to this was outside of the scope of this study, this lends support to the explanation that news events occurring during elections affect online discussions. 

In addition to Reddit, \citet{zahrah} has conducted a comparative study on 4chan against the former, analyzing hateful content related to the 2020 US Presidential Elections from the two platforms. Through this, they suggested that different platforms can serve specific purposes. Twitter, as a microblogging website, is useful to gain insight into the immediate reactions of individuals to events. 
While Twitter was found to be effective for election predictions by \citet{Tumasjan}, \citet{bryant} found that sentiments correlated with election results, though sentiment analysis for elections on this platform is not as robust as that of Twitter. For example, the US elections have electoral colleges, where peoples' votes weigh differently according to which state the vote comes from; without location data on Reddit, researchers may find that the accuracy will fall. In the case of this study, since the target was the Japanese House of Councillors elections, the location was not as relevant. However, Reddit was not selected due to it not being as popular of a platform as Twitter in Japan with only a limited number of Japanese-language subreddits.

While not related to elections, \citet{melton} has used a similar methodology as this study in their research on public sentiments towards COVID-19 vaccines on Reddit. Their data was aggregated and analyzed by month while for this research, data were aggregated and analyzed by day. Instead of looking at positive scores between 0 to 1, they utilized polarity analysis that ranges from -1 to 1. While no emotion analysis was conducted, Latent Dirichlet Allocation (LDA) topic modeling with Gensim was also conducted. This research referred to their methodology for LDA topic modeling for conducting topic modeling on Twitter sentiment, though there were some changes made in how data was lemmatized and processed due to the language of the original data being different - the COVID-19 study's collected data was in English while this study's data was in Japanese. 

\subsection{Martyrdom and its Politicization}

This section will now examine related works on necropolitics and political martyrdom. Studies by many scholars across the globe have examined the political implications of martyrdom and their remembrance. While \citet{mbembe2008necropolitics} asserts that "sovereignty resides, to a large degree, in the power and the capacity to dictate who may live and who must die," in the context of this research, we will focus instead on the aftermath of the removal of a former sovereign figure.

When one first hears of martyrs, the first impression might be of religious martyrdom. An example of which is \textit{jihad} which originates in the Middle East. 
As has been discussed by \citet{Allen}, people, especially in Middle Eastern nations, can be united through the remembrance of tragedy through the practice of commemorating martyrs. In times of political instability or change, this can serve as a powerful factor in gaining support from the people. For example, in the case of Iran, the popularization and politicization of martyrdom in the 1960s have functioned to unite people, enhance mass mobilization, and preserve tradition \citep{dorraj}. While not to this extent, the government's treatment of former Prime Minister Shinzo Abe's death may have mobilized people to conduct pity votes for Abe's Liberal Democratic Party (LDP), a conservative political party that is the largest in Japan and since its establishment, has been in power almost consecutively \citep{ldp}. With regards to jihad, \citet{euben} suggests that rather than death killing politics, death may instead be the precondition to politics.

Meanwhile, in China, where there is a deep history of martyrdom, \citep{hung} explored the political implications of red martyr cults, where the Chinese Communist Party (CCP) attempted but failed to completely impose their political agenda on the local martyrs in the 1950s. The failure was partly attributed to the private nature of local deaths for the bereaved and strong local politics that overshadowed that of the state. In a similar manner, in the Roman Empire, provincial understandings eclipsed that of the imperial government, resulting in lasting effects of the provincial-imperial relations that have remained in today's scholarly models \citep{bryen}. In the case of Abe, this may not be as significant of an obstacle, due to Abe has been a major public figure and a former representative of the state.

In general, \citet{desoucey} suggests that martyrs are tangible cultural resources that assist in the understanding of the societal collective, where their death and legacy calls for some form of action, where examples raised were Joan of Arc and Che Guevara, and demonstrated by \citet{hyder2006reliving} through their exploration of the death of the grandson of the Prophet Muhammad, Husain b. Ali, at Karbala. For Abe's death, depending on how the Japanese government handles the assassination of Abe, it may result in Abe's long-term embodiment of a martyr figure. While this is outside the scope of this study, results observed in this study are intended to provide insight into the online sphere of election sentiments and much effect Abe has provided in the past and may provide in future House of Councillors elections.

\section{Method}

The code for data collection has been made publicly available in the interest of transparent open science (GitHub.com/mint-talltree/Abe-Assassination-Election-Tweet). 

While the main target of analysis is data collected from 2022, data from elections preceding that were also collected to have historical data to compare against to best observe the effects Abe’s assassination had if there were any. As mentioned by \citet{Jungherr}, “we can distinguish between two approaches: one, relying on scripts developed by researchers querying Twitter’s API or scraping Twitter’s Web site, the other, using third-party software for data collection on Twitter." This paper has taken the first approach instead of the latter, to gain more control in data collection to customize the data collection date, language of tweets, and key terms mentioned in the tweet. As such, tweets in a Twitter thread that did not include the key terms in their text were excluded.

Tweets in Japanese were collected from the years 2007, 2010, 2013, 2016, 2019, and 2022. 

More specifically, the period of collection for each year was 18 days immediately before Election Day. The period of 18 days was selected due to there being 18 days before the 2022 Election Day since when the elections were first announced to the public; in other words, for example, the days for 2022 ranged from June 22nd to July 9th. In the case of 2013 and 2019, the date range for data collection was July 3rd to July 20th. Similar to 2022, for 2016, the dates ranged from June 22nd to July 9th. While the election for 2007 was originally slated to be conducted on July 22nd, it was postponed by a week in mid-June to July 29th. As such, the range of dates was July 11th to July 28th. Finally, for 2010, since the elections were held on July 11th, tweets posted between June 23rd to July 10th were collected. 

Tweets were collected based on the criteria that they were Japanese-language tweets that contained the words “
\begin{CJK}{UTF8}{min}
選挙
\end{CJK}
” (election), “
\begin{CJK}{UTF8}{min}
参院選
\end{CJK}
” (House of Councillors Election), “
\begin{CJK}{UTF8}{min}
参議院選挙
\end{CJK}
” (long form of House of Councillors Election), “
\begin{CJK}{UTF8}{min}
選挙
\end{CJK}
Year Number” (Election Year Number), “
\begin{CJK}{UTF8}{min}
参院選
\end{CJK}
Year Number” (House of Councillors Election Year Number), or “
\begin{CJK}{UTF8}{min}
参議院選挙
\end{CJK}
Year Number” (long form of House of Councillors Election Year Number), with the respective years typed in the place of Year Number. Examples for the last three key terms are the following:
"\begin{CJK}{UTF8}{min} 
選挙2022
\end{CJK}
" (Election 2022), "
\begin{CJK}{UTF8}{min} 
参院選2022
\end{CJK}
" (House of Councillors 2022), and "
\begin{CJK}{UTF8}{min} 
参議院選挙2022
\end{CJK}
" (long form of House of Councillors 2022).

Including keywords such as “
\begin{CJK}{UTF8}{min}
安部
\end{CJK}
” (Abe), “
\begin{CJK}{UTF8}{min}
暗殺
\end{CJK}
” (assassination), and “
\begin{CJK}{UTF8}{min}
射殺
\end{CJK}
” (shot dead) were initially considered but were rejected in order to capture the election sentiments as a whole, rather than tweets specifically geared towards Abe and the election, in addition to comparisons of 2022 tweet sentiments against those from 2007, 2010, 2013, and 2016 being planned. Before pre-processing of tweets, there were 786 tweets for 2007, 262,298 tweets for 2010, 800,033 tweets for 2013, 1,003,129 tweets for 2016, 932,054 tweets for 2019, and 2,060,631 tweets for 2022 collected. 


In order to process the Japanese text, MeCab \citep{MeCab}, a morphological analyzer engine, was used in combination with fugashi \citep{fugashi}, a Cython wrapper for MeCab, with the UniDic-lite dictionary installed. Data were cleaned using regex to remove websites and superfluous characters from the text. Examples of superfluous characters removed were punctuation marks such as "!" and """". In addition, duplicate tweets were removed to condense the data down to 561 tweets for 2007, 213,426 tweets for 2010, 645,839 tweets for 2013, 790,483 tweets for 2016, 814,975 tweets for 2019, and 1,432,705 tweets for 2022.

Following the cleaning, tweets were put through the Tokenizer from janome.tokenizer to tokenize the text for sentiment analysis.

Tweets were first classified into positive and negative sentiments. For the classification of tweets into positive and negative sentiments with confidence scores ranging from 0 to 1, a pre-trained BERT model, "daigo/bert-base-japanese-sentiment" \citep{daigo}, was utilized. 
While training a positive/negative classifier on WRIME Ver.2 data \citep{wrime}, which, compared to Ver. 1, contains an additional column of Emotion Polarity scores between -2 and 2 were considered, this method was rejected due to unknown crashes of the computer seemingly correlated to the running of the code to train and fine-tune a model. Following the classification, tweets with negative sentiments were processed to subtract their scores from 1 to get the confidence scores in relation to positive sentiments. 

\begin{figure}[ht]
    \centering
    \includegraphics[scale=0.4]{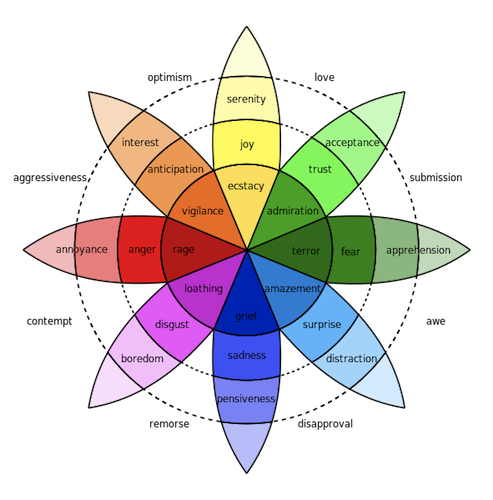}
    \caption{Plutchik's Wheel of Emotions}
    \label{fig:plutemot}
\end{figure}

\begin{figure}[ht]
    \centering
    \includegraphics[scale=0.5]{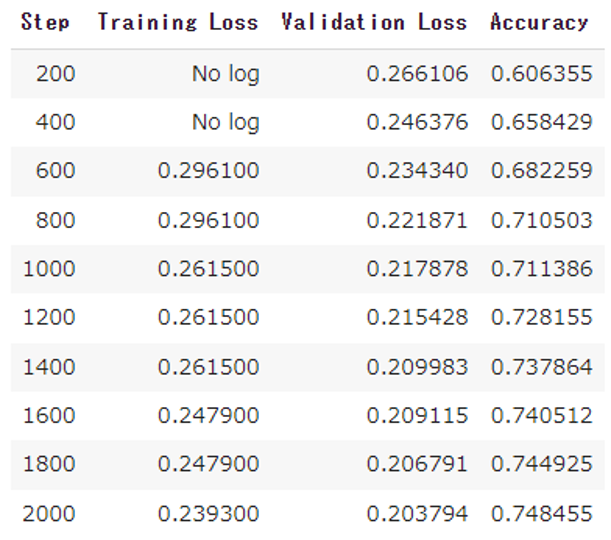}
    \caption{Validation Scores for BERT}
    \label{fig:berttrain}
\end{figure}

In addition to positive and negative sentiments, tweets were classified into Plutchik's 8 basic emotions \citep{plutchik}, illustrated in Figure \ref{fig:plutemot}; happiness, sadness, anticipation, surprise, anger, fear, disgust, and trust. For each tweet, the tweet had proportions allocated to each emotion such that when the 8 emotions were added up for each tweet, they would equal 1. In order to create the classifier, the same pre-trained model, "daigo/bert-base-japanese-sentiment" \citep{daigo}, was utilized. To train the classifier, Trainer Huggingace Transformers was employed \citep{HuggingFace}, with validation scores printed out, as seen in Figure \ref{fig:berttrain} to ensure that the machine was improving every 200 steps. As the training proceeded, accuracy increased, while both training and validation losses decreased, showing the improvement of the machine, though improvement seemed to approach a plateau near the end. From WRIME data, columns starting with ‘Avg. Readers
\textunderscore
’ were used to train the machine. The checkpoint was set to 'cl-tohoku/bert-base-japanese-whole-word-masking.'

To analyze the data, the positive and negative sentiments were first plotted against date. In order to minimize the noise, sentiments were aggregated by day. In other words, they were grouped by day and averages for each day was calculated using groupby. Emotion sentiments were processed and plotted in a similar fashion. 2022 polar sentiments were compared against 2007, 2010, 2013, 2016, and 2019 data in order to gain insight into whether Abe’s assassination had an effect on election sentiments. 

Since former Prime Minister Shinzo Abe belonged to the Liberal Democratic Party (LDP), the effects of his assassination on election tweets mentioning the LDP, “
\begin{CJK}{UTF8}{min}
自民党
\end{CJK}
” (abbreviation for the LDP) and “
\begin{CJK}{UTF8}{min}
自由民主党
\end{CJK}
” (Liberal Democratic Party), were extracted from the processed 2022 tweets. To visualize diagrams related to the Liberal Democratic Party, the party Abe belonged to, BERT and WRIME sentiments were plotted for the entire period, in a similar manner to those listed above. Additional plots of the same sentiments were also created for the period between July 7th to July 9 to see more clearly the changes in sentiment if there were any. 

\begin{figure}
\centering     
\subfigure[2022 Overall]{\label{fig:coherenceoverall}\includegraphics[width=49mm]{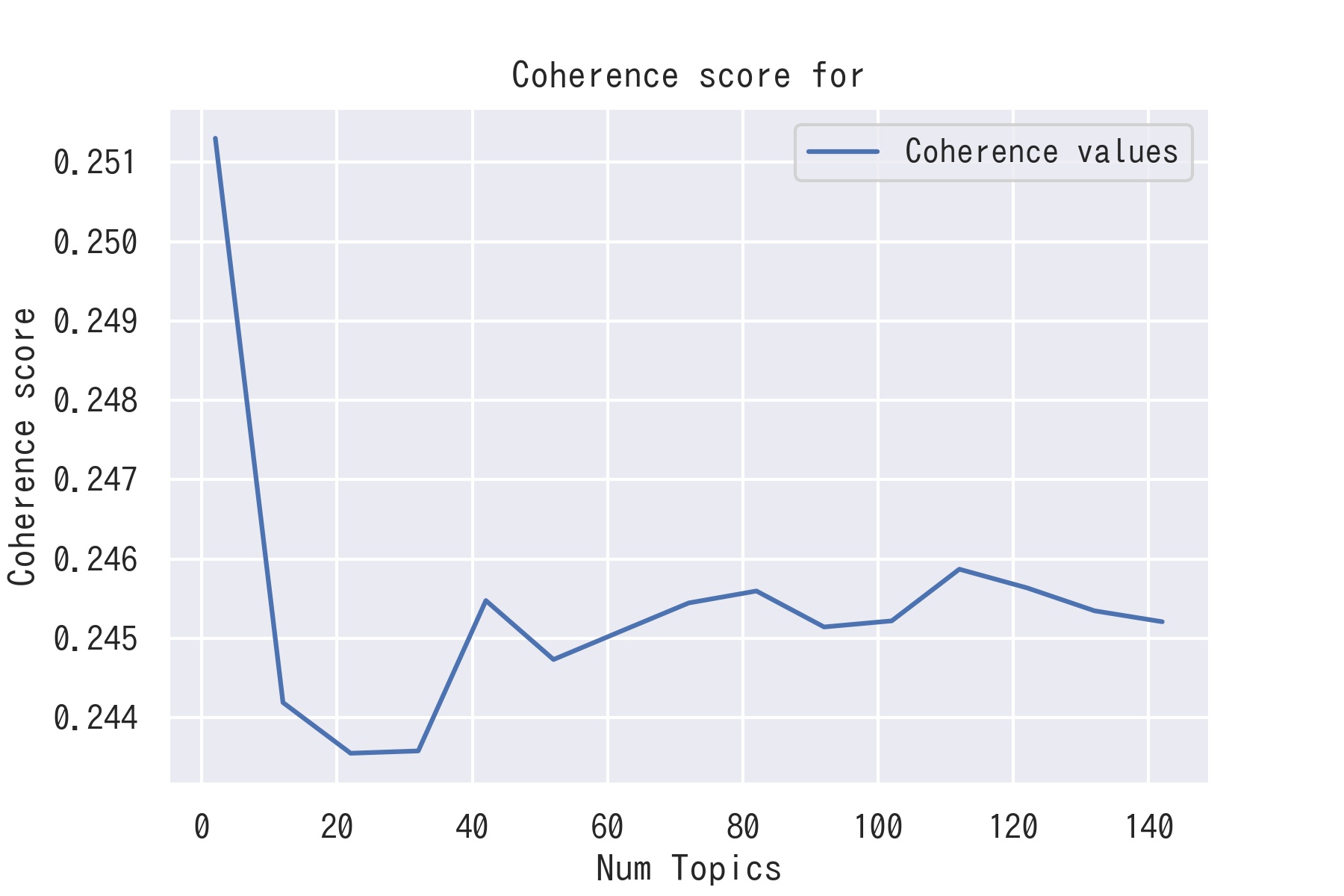}}
\subfigure[Before Assassination]{\label{fig:coherencebefore}\includegraphics[width=49mm]{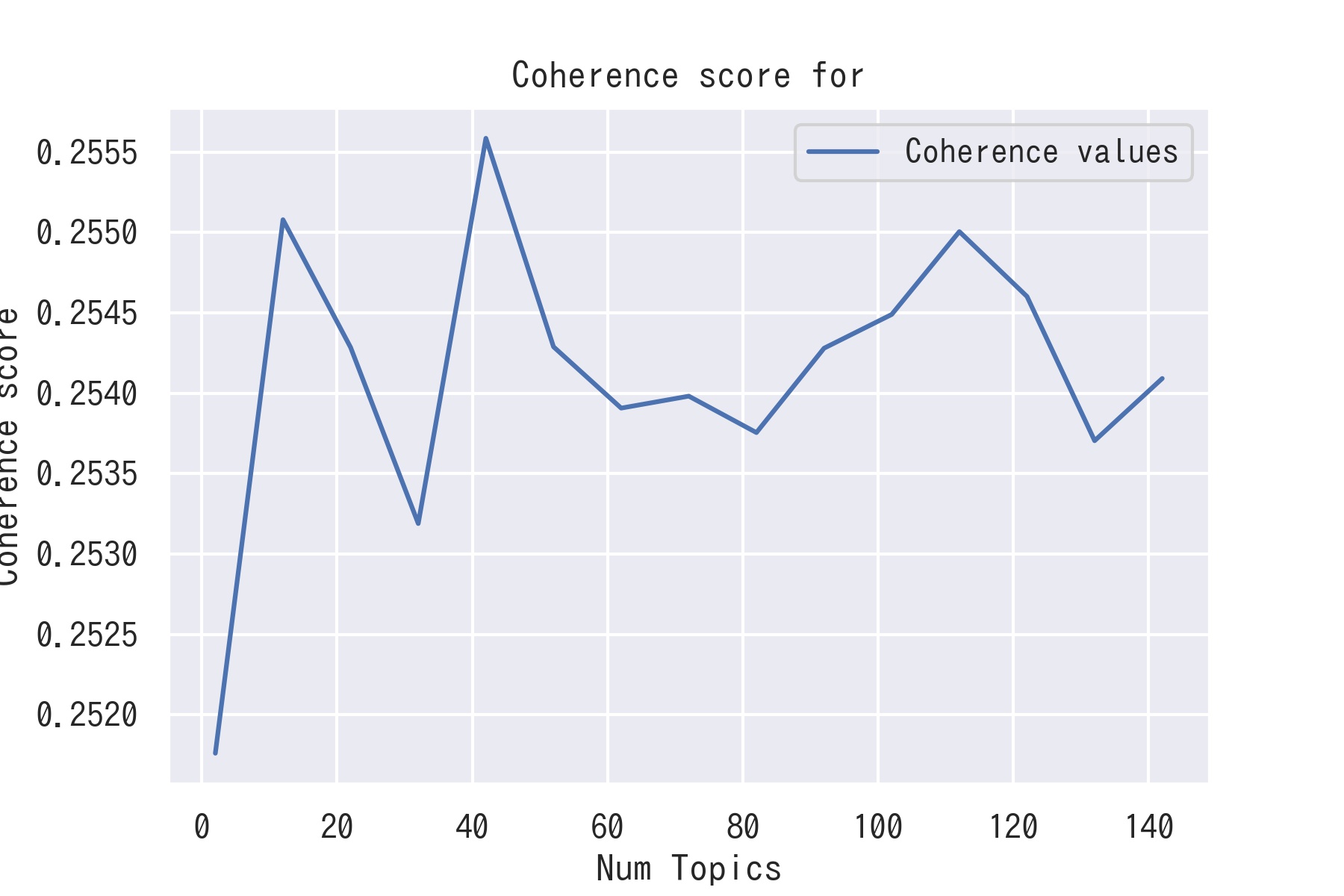}}
\subfigure[After Assassination]{\label{fig:coherenceafter}\includegraphics[width=49mm]{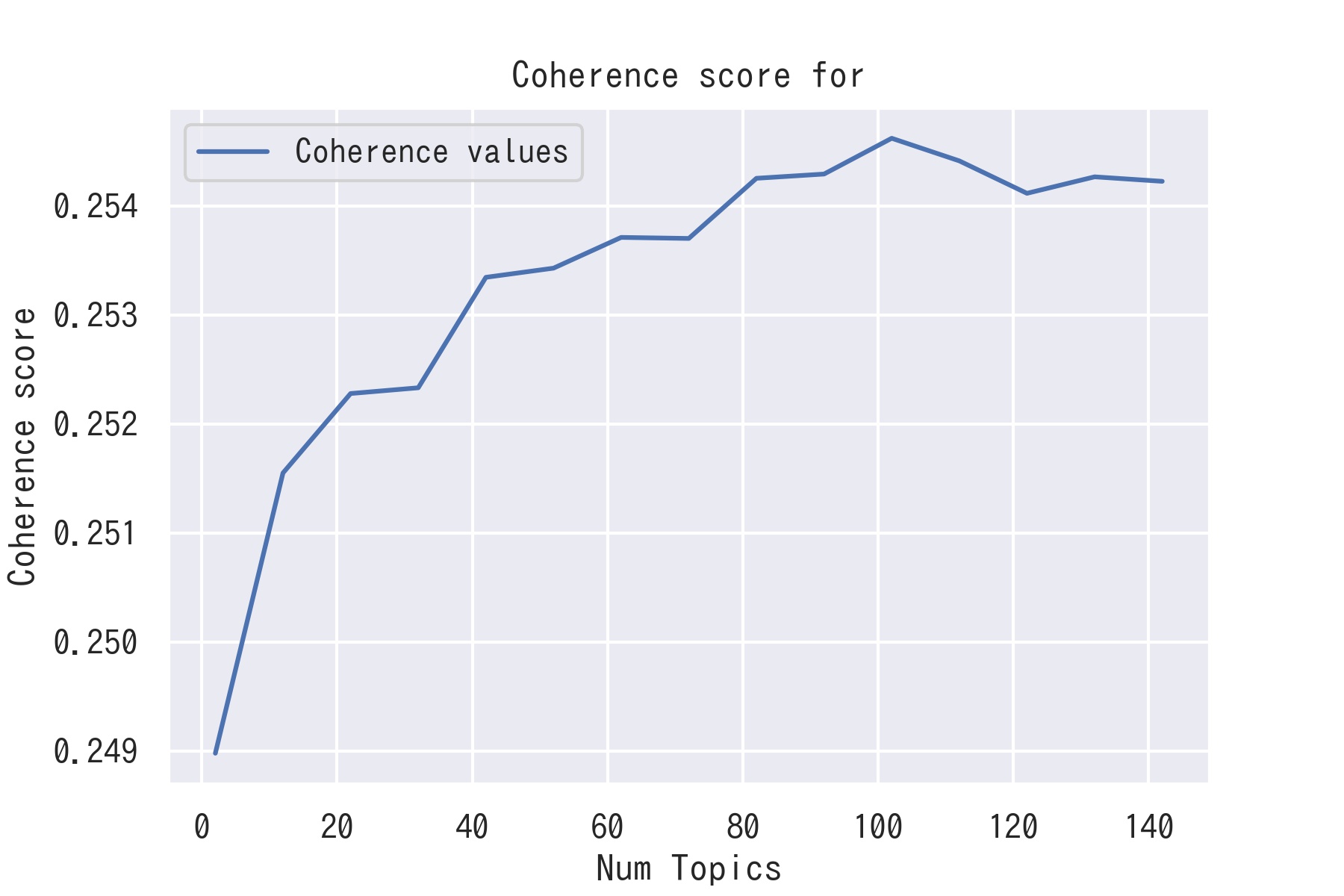}}
\caption{Coherence Scores for Topic Modelling}
\end{figure}

Additionally, Topic Modeling was conducted for 2022 data in order to view popular topics discussed related to the elections to further extract the effect Abe’s passing had. SlothLib stopwords were downloaded and expanded to include punctuation marks, standalone “―,” and “
\begin{CJK}{UTF8}{min}
選挙
\end{CJK}
” (election), “
\begin{CJK}{UTF8}{min}
参院
\end{CJK}
” (abbreviation for House of Councillors), and “
\begin{CJK}{UTF8}{min}
参院選
\end{CJK}
” (abbreviation for House of Councillors Election). The full form of words was not included due to them being scarcely used in tweets, and not being outputted during the first round of Topic Modeling without the extended stopword dictionary. SlothLib \citep{oshima2007slothlib} was selected for use due to it being specifically created as a programming library for research on the Web. Tokenized tweets from the previous BERT training for the 8 emotions were reused. Bigram and Trigram models were next created with spacy. Bigrams were made and data was lemmatized. A dictionary and corpus were then created before building the topic model. The topics were set to 42 to build the topic model, following an evaluation of coherence scores after iterating through to build several LDA models starting from 2 topics with steps of 10, to which the program was set to terminate when it hit a limit of 100 topics. The coherence scores can be observed in Figures \ref{fig:coherenceoverall}, \ref{fig:coherencebefore}, and\ref{fig:coherenceafter}.

\section{Results}

 \subsection{Word distributions}

For Tweets from the 2007 Japanese House of Councillors election, the most frequent words can be observed to be 
\begin{CJK}{UTF8}{min}
投票
\end{CJK}
(vote), 
\begin{CJK}{UTF8}{min}
明日
\end{CJK}
(tomorrow), 
\begin{CJK}{UTF8}{min}
今日
\end{CJK}
(today), 
\begin{CJK}{UTF8}{min}
カー
\end{CJK}
(unknown), and 
\begin{CJK}{UTF8}{min}
候補
\end{CJK}
(candidate). In 2010, the topics seem to shift slightly to involve 
\begin{CJK}{UTF8}{min}
こと
\end{CJK}
(thing) and 
\begin{CJK}{UTF8}{min}
消費税
\end{CJK}
(tax). 2013 saw the rise of 
\begin{CJK}{UTF8}{min}
ネット
\end{CJK}
(internet), 
\begin{CJK}{UTF8}{min}
山本太郎
\end{CJK}
(Taro Yamamoto), and 
\begin{CJK}{UTF8}{min}
期日前
\end{CJK}
(before deadline) while 2016 observed 
\begin{CJK}{UTF8}{min}
三宅洋平
\end{CJK}
(Yohei Miyake) and 
\begin{CJK}{UTF8}{min}
都知事
\end{CJK}
(city mayor). 2019 mainly involved the key terms 
\begin{CJK}{UTF8}{min}
前投票
\end{CJK}
(early voting), 
\begin{CJK}{UTF8}{min}
期日前
\end{CJK}
(before deadline), 
\begin{CJK}{UTF8}{min}
新選組
\end{CJK}
(Shinsengumi Party), and the re-rise of 
\begin{CJK}{UTF8}{min}
山本太郎
\end{CJK}
(Taro Yamamoto). Popular key terms in 2022 overall were 
\begin{CJK}{UTF8}{min}
前投票
\end{CJK}
(early voting), 
\begin{CJK}{UTF8}{min}
期日前
\end{CJK}
(before deadline), and 
\begin{CJK}{UTF8}{min}
民主主義
\end{CJK}
(democracy). 

There was one common key term between most Japanese House of Councillors elections studied; 
\begin{CJK}{UTF8}{min}
期日前
\end{CJK}
(before deadline) was observed for years 2013, 2016, 2019, and 2022.

Comparing key terms before and after the assassination of former Prime Minister Abe, one can observe that while the key terms 
\begin{CJK}{UTF8}{min}
前投票
\end{CJK}
(early voting) and 
\begin{CJK}{UTF8}{min}
期日前
\end{CJK}
(before deadline) are common, following the assassination, 
\begin{CJK}{UTF8}{min}
民主主義
\end{CJK}
(democracy), 
\begin{CJK}{UTF8}{min}
明日
\end{CJK}
(tomorrow), and 
\begin{CJK}{UTF8}{min}
こと
\end{CJK}
(thing) were popular key terms to tweet, along with key terms related to 
\begin{CJK}{UTF8}{min}
安部
\end{CJK}
(Abe).

\subsection{Time Series}
\label{timeseries}
\begin{figure}[ht]
    \centering
    \includegraphics[width=\columnwidth]{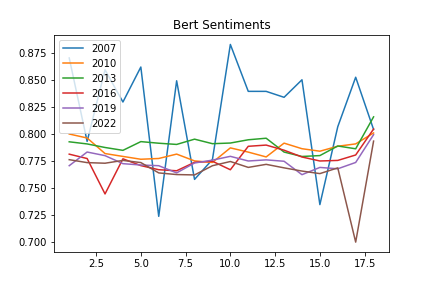}
    \caption{BERT Sentiments}
    \label{fig:bertsent}
\end{figure}

Averaged tweet sentiments for each observed day were plotted out for each year in \ref{fig:bertsent} to observe the changes in sentiment for each day of the election period. 

In 2007, sentiments could be seen to jump largely between a confidence score of 0.875 and 0.725 quite unpredictably. Sentiments for this year were at an average of 0.820317, though most lay above 0.800.

For all years, with the exception of 2007, 2016, and 2022, sentiments followed a similar trend. Sentiments generally stayed stable in the range of 0.800 and 0.7625 from days 1 to 17, after which sentiments have a small uptick to the range of 0.775 to 0.825. 2016 and 2022 follow a similar trend, though, for 2015, there is a temporary fall to 0.750 in positive sentiment on day 3, while for 2022, day 17 observed a great temporary drop in sentiment to 0.700, the lowest recorded positive sentiment score. 

For all years, sentiments settled in the range of 0.775 and 0.825, lying close to 0.800 on the 18th day of the election period, the day before voting day. 

\subsection{2022 Elections}
\label{2022elec}
Focusing on 2022 Bert sentiments alone in \ref{fig:bertsent}, while sentiments were stable from days 1 to 16, sentiments seem to have gradually decreased over time before taking a plunge on day 17. Sentiments however recovered on day 18. 

\begin{figure}[ht]
    \centering
    \includegraphics[width=\columnwidth]{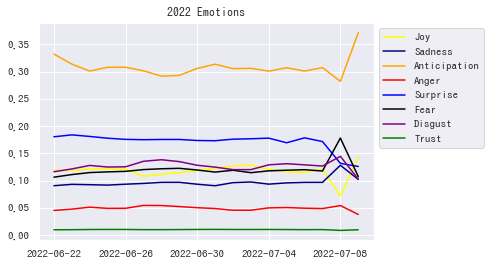}
    \caption{Plutchik's 8 Emotions for 2022}
    \label{fig:wrimesent}
\end{figure}

Digging further into the 2022 sentiments, in \ref{fig:wrimesent} showing the sentiments divided into Plutchik’s 8 emotions of joy, sadness, anticipation, anger, surprise, fear, disgust, and trust, on can observe that the feeling of “Trust” was negligible throughout the 2022 House of Councillors elections. Meanwhile, “Anticipation” was the strongest emotion, generally lying in the range of 0.30 to 0.35 throughout the observed period. “Surprise” was the second strongest sentiment in general, falling in the range of 0.15 to 0.20 from June 22 to July 7, after which the score fell to 0.10 to 0.15, indicating a fall in the amount of “Surprise” observed through tweets. The emotions “disgust” and “anger” stayed relatively stable throughout the observation period, in the range of 0.10 to 0.15 and 0.05 respectively. “Sadness” and “Fear” also stayed relatively stable between June 22 and July 7, with “Fear” generally scoring slightly higher than “Sadness.” On the day of Former Prime Minister Abe’s assassination, sentiment scores rose to 0.178147 for “Fear” and 0.128134 for “sadness”, after which scores fell approximately between 0.10 and 0.11 on July 9. Finally, much like the other sentiments, “Joy” was observed to be relatively stable in the range of 0.10 and 0.15 during the first 16 days. On day 17, “Joy” fell to a score of 0.072420, after which it recovered to 0.142329 on the final day, a pattern opposite that of “Fear” and “Sadness.”

Notable differences between the trigrams for 2022 Japanese House of Councillors elections overall in \ref{fig:2022 Overall Trigram}, before the assassination in \ref{fig:Before Assassination}, and after in \ref{fig:After Assassination}. Overall, the top three topics were "
\begin{CJK}{UTF8}{min}
わたし
\end{CJK}
" (me), “
\begin{CJK}{UTF8}{min}
考え
\end{CJK}
” (thought), “
\begin{CJK}{UTF8}{min}
政治
\end{CJK}
” (politics), [“2022.” “
\begin{CJK}{UTF8}{min}
わたし
\end{CJK}
” (me), “
\begin{CJK}{UTF8}{min}
考え
\end{CJK}
” (thought)], and [“
\begin{CJK}{UTF8}{min}
選挙ドットコム
\end{CJK}
” (Election dot com), “
\begin{CJK}{UTF8}{min}
参院選
\end{CJK}
” (House of Councillors election), “
\begin{CJK}{UTF8}{min}
投票
\end{CJK}
” (vote)]. During the period before the assassination, the top 5 topics were an exact match to that of 2022, with topics regarding quizzes on who one should vote for, 
\begin{CJK}{UTF8}{min}
ボートマッチ
\end{CJK}
(vote match) being the most popular topic tweeted. Following the assassination, however, most popular topics are no longer about which party one’s political preference is closest to, but about [“
\begin{CJK}{UTF8}{min}
仁藤
\end{CJK}
” (Nito), “
\begin{CJK}{UTF8}{min}
夢乃
\end{CJK}
” (Yumeno), “
\begin{CJK}{UTF8}{min}
氏
\end{CJK}
” (mister)], [“
\begin{CJK}{UTF8}{min}
安部政治
\end{CJK}
” (Abe politics), “
\begin{CJK}{UTF8}{min}
要因
\end{CJK}
” (cause/factor), “
\begin{CJK}{UTF8}{min}
主張
\end{CJK}
” (claim)], [“
\begin{CJK}{UTF8}{min}
安部元首相
\end{CJK}
” (former Prime Minister Abe), “
\begin{CJK}{UTF8}{min}
射殺
\end{CJK}
” (shot dead), “
\begin{CJK}{UTF8}{min}
安部政治
\end{CJK}
” (Abe politics)], [“
\begin{CJK}{UTF8}{min}
氏
\end{CJK}
” (mister), “
\begin{CJK}{UTF8}{min}
安部元首相
\end{CJK}
” (former Prime Minister Abe), “
\begin{CJK}{UTF8}{min}
射殺
\end{CJK}
” (shot dead)], and [“
\begin{CJK}{UTF8}{min}
女性の権利
\end{CJK}
” (women’s rights), “
\begin{CJK}{UTF8}{min}
軽視
\end{CJK}
” (neglect/slight), “
\begin{CJK}{UTF8}{min}
怒り
\end{CJK}
” (anger)]. Separate from Abe’s assassination, the topic of politician Yumeno and women’s rights seemed to be popularly tweeted about. Tweets regarding vote matching fell to 6th place and below.

More detailed results are in the appendix.
\section{Analysis and Discussion}
\subsection{Sentiments Overall}
To answer hypothesis 1, overall, former Japanese Prime Minister Shinzo Abe's assassination seems to have prominently affected Twitter election sentiments negatively on the short term.

In general, positive/negative sentiment ratios seem to be the same for all Japanese House of Councillors elections. While there may be fears that Twitter is an echo chamber that is taken advantage of, especially by “Trolls” and “Bots,” observing the general stability of sentiments, with sentiments generally remaining positive rather than negative, it could perhaps be said that Japanese Twitter during elections, in general, seem to not be that affected by “Trolls” and “Bots,” and are not as polarized as elections in places, such as the US. 

While it is not surprising that “Anticipation” was not affected by the assassination of Shinzo Abe, “Trust” being quite low and not being affected at all may have been surprising for some. It should, however, be noted that trust is quite difficult to detect in text. Especially, amount of trust is generally detected by people through keywords such as “trust” or “lie.” As such, “Trust” may not be a reliable indicator in the case of tweet sentiment analysis for elections and could most likely be safely ignored.
While the average amount of surprise felt fell, this could be explained by the amount of fear and sadness overtaking the other emotions, resulting in there being less surprise proportionally, though the event was sudden and unexpected. 

\begin{figure}[ht]
    \centering
    \includegraphics[scale=0.5]{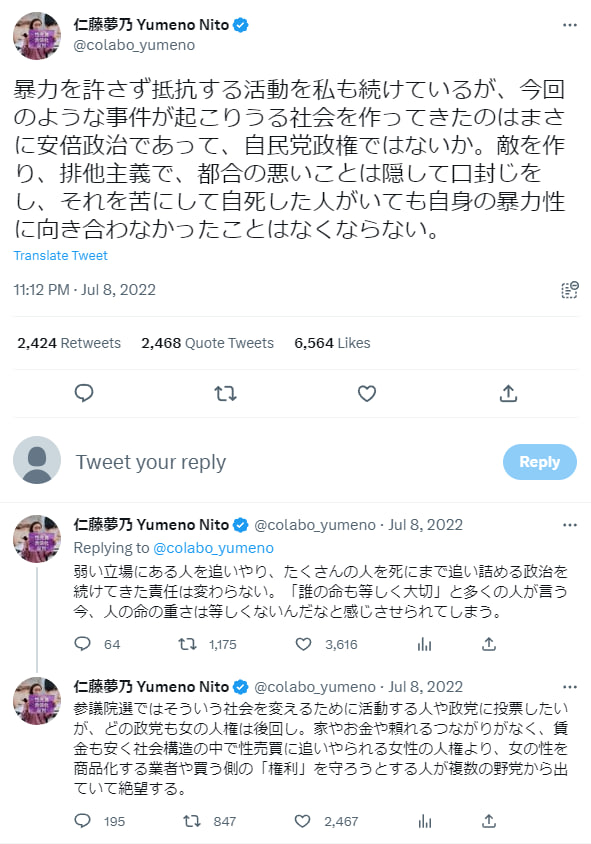}
    \caption{Yumeno Nito's tweet Reacting to Abe's Assassination}
    \label{fig:yumeno}
\end{figure}

The topics having topics unrelated to former Prime Minister Abe being popular following the assassination were needless to say, unexpected. In fact, it scoring the highest in terms of topics was surprising. Meanwhile, on the news, this news did not come much to the attention of the researcher, with almost all news observed on television being about the accomplishments of former Prime Minister Shinzo Abe, and the situation during the assassination. Following the assassination, the candidate involved, Yumeno Nitou, stirred the discussion on Twitter, tweeting that
\begin{CJK}{UTF8}{min}
今回のような事件が起こりうる社会を作ってきたのはまさに安倍政治であって、自民党政権ではないか
\end{CJK}
(it is precisely Abe's politics and the Liberal Democratic Party (LDP) government that has created a society where incidents like this can occur), as seen in 
Figure \ref{fig:yumeno}. Since she had also mentioned that 
\begin{CJK}{UTF8}{min}
参議院選ではそういう社会を変えるために活動する人や政党に投票したいが、どの政党も女の人権は後回し
\end{CJK}
(Though I want to vote for people and parties that try to work towards changing such societies in the House of Councillors election, every party does not prioritize women's rights). This reaction may be part of the reason why there was a rise in topic popularity. Additionally, at the time there seemed to be online conflict against game developer Akane Himazora, on women's rights. As diving into this topic digresses from the main point of discussion, the effects of Abe's assassination on Twitter's landscape, and enters the realm of politics, further exploration, and discussion into her rise in topic popularity will be avoided. 

Overall from Figure \ref{fig:wrimesent} showing 2022 election emotion proportions and Figure \ref{fig:bertsent} showing 2022 Bert positive sentiment proportion over time, it can be seen that Abe’s passing only had a momentary effect on tweets containing election-related words tweeted. As such, while Abe’s assassination may have had some impact on the electoral results, the event could be said to have had a very short, temporary effect that seemingly did not result in any long-term impact. 

As such, while former Prime Minister Shinzo Abe may have been a very impactful figure during his time as Prime Minister, and cast a large shadow over his own Liberal Democratic Party, it is doubtful to say that his passing had any lasting, permanent impact on future elections. 

\subsection{Social Media Attention Span}
While not directly related to Abe’s assassination and its effects, the results of this study indicate a fascinating change in the public’s attention span. Attention span here will be measured as the difference in time taken from the start of notable change in sentiment observed to recovery, in hours. Plutchik’s 8 emotions were not included due to this processing only being conducted for 2022 data. 

\begin{figure}[ht]
    \centering
    \includegraphics[scale=0.8]{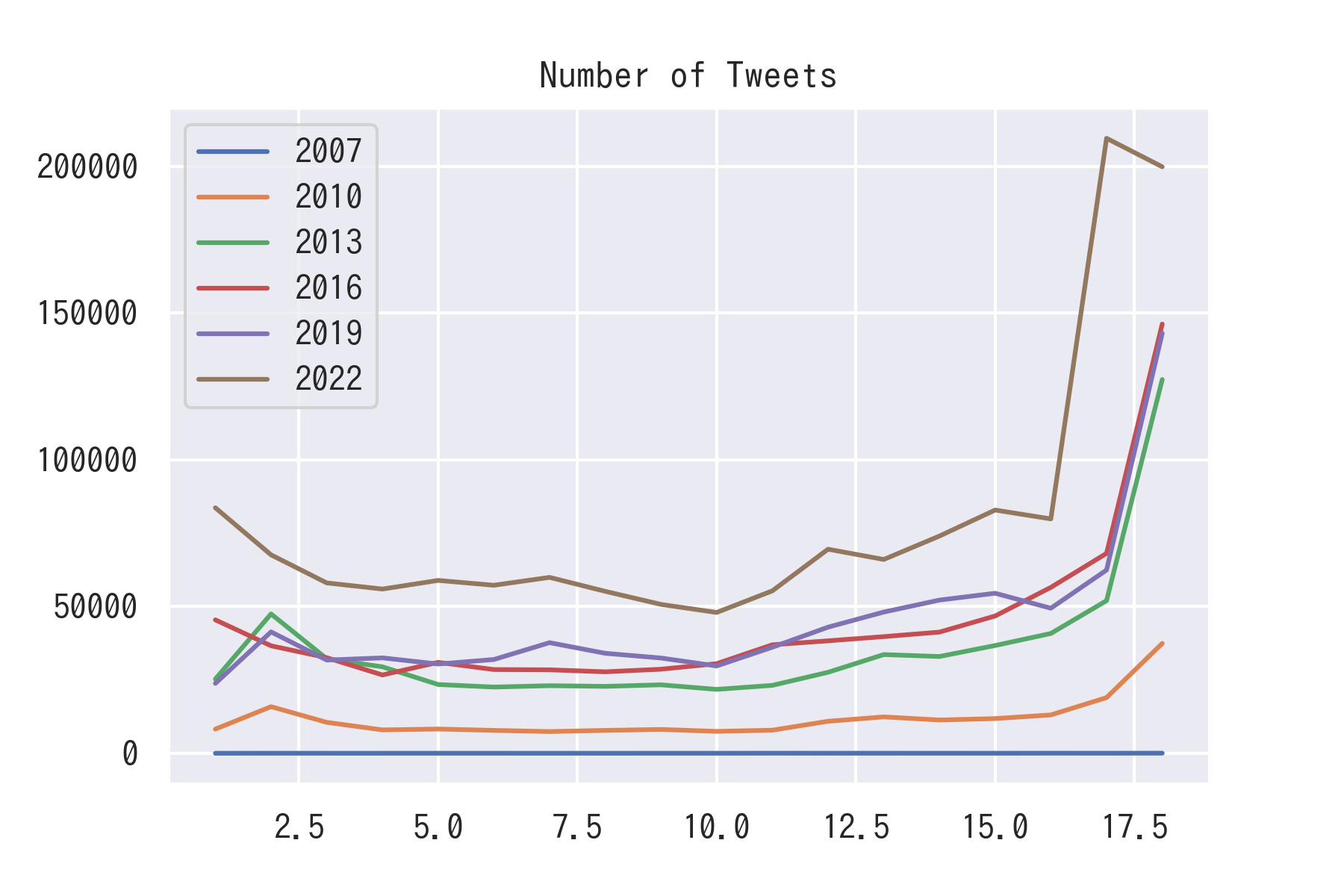}
    \caption{Number of tweets per Day}
    \label{fig:occurrences}
\end{figure}

As indicated in this study, the public’s attention seems to have shortened to just a few hours that does not even last an entire day, though this may be because elections are not directly related to assassinations. However as seen in Figure \ref{fig:occurrences}, the number of tweets overall greatly increased on the day of the assassination. As such, this study may provide some validity in the exploration of social media attention span length changes. 

With an increasingly shorter amount of time spared in the attention given to a piece of fresh news, alongside the rise of social media platforms showcasing shorter attention- grabbing videos, it puts into question whether this is a real problem or if society has just become better at managing feelings of crisis. 

This phenomenon of short attention spans seems to have already been observed in 2016. 2016 is thought to have observed a drop in sentiments on the 3rd day due to the EU Brexit results being released that day, with the UK votes leaning towards leaving the EU. This is thought to have a temporary dampening effect on Japanese election sentiments, though Brexit was less tweeted about compared to other topics overall. 

Given that the public’s attention span seems to have already been shortened to a day by 2016, 2022 may have an even shorter attention span, though the type of news and how close to home the event is, is quite different, though they are both shocking no less. 
As such, sentiments during those three days were plotted hourly to see whether this second hypothesis is correct.

Compared to 2016, 2022’s sentiments recovered approximately 18 hours faster, at 19 hours. This may imply that the attention span and trend lifetime of news on social media, especially with regard to those impacting elections, have declined. While outside the scope of this paper, further research may be necessary to compare attention span lengths between various topics. Depending on the results, there may be findings on how the public processes news and how impactful news are on the public consciousness.

Looking deeper at the data, the news of Abe’s death seems to have spread slower than the news of his being shot earlier the same day. For tweets containing the key term of LDP in the figure below, sentiments fell to a similar level as when news circulated for election sentiments overall, though sentiments towards the LDP were on a falling trend throughout the collection period. While the positive-negative sentiment seems to have recovered, Plutchik's 8 emotions did not seem to recover following the assassination. This is most likely a result and reflection of Abe wielding much clout within the LDP, so much so that he was known as the kingmaker in 2021 \citep{Harris}. An additional explanation for the faster recovery for election sentiments overall could be that the gun being shot in public had a larger impact on the Twitter community, given that the key term for being shot to death ranked quite high in the topic modeling. Further research into sentiments for specific key terms was outside the focus of this study on Abe's impact.
\begin{figure}
\centering     
\subfigure[WRIME Sentiments]{\label{fig:ldpWRIME}\includegraphics[width=70mm]{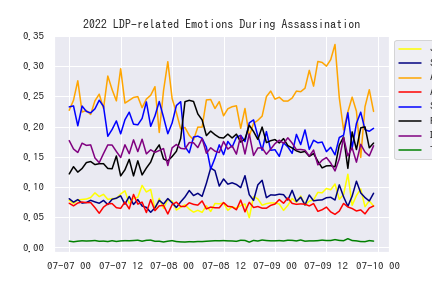}}
\subfigure[BERT Sentiments for LDP]{\label{fig:ldpbert}\includegraphics[width=70mm]{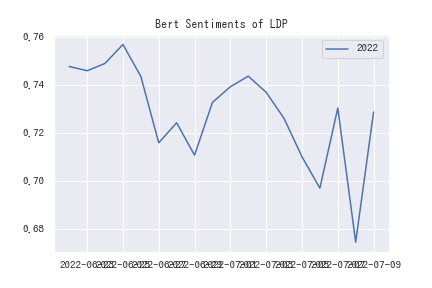}  }

\caption{Coherence Scores for Topic Modelling}
\end{figure}

Of course, this study was not conducted to diminish the accomplishments of the late former Prime Minister Shinzo Abe, nor was it done to criticize social media. Nevertheless, with there being no end to the list of long-term problems and issues; how to maintain public interest by keeping news fresh, is a concerning matter. Especially with time-sensitive issues, such as climate change, reporting the news from different angles or aspects may be integral in keeping news seemingly fresh. At the same time, care should be taken so that the distributors of news are not seen as being tone-deaf or needlessly dragging certain issues on to “milk” the attention of viewers. Especially with the government handling of Abe’s funeral, criticisms were raised on the “tone-deafness” of the government and their “over-evaluation” of the impact of Abe. In comparison to the short social media attention span, news reporting on the assassination of Abe has been observed to continue for days with regard to the election and months on just the assassination alone. Additionally, the government’s attention with regard to the assassination was observed to continue for months as well. In fact, there were some tweets criticizing that the government was attempting to turn Abe into a martyr to shift the public’s view from other issues such as Abe and government officials’ involvement with cults. 

\subsection{Necropolitics}
With hypothesis 2, while Abe's assassination indicate that deaths may sway electoral votes, no definitive conclusion was drawn on the event setting a precedent for necropolitics.

Before the review on \textit{necropolitics}, however, it is first important to explore what martyrdom is. As \citet{middleton} has found, the outer bounds of the definition of martyrs and martyrdom have been elusive. There have been debates on whose deaths to include within the definition. Especially, Middleton argues that martyrology can be created separately from the actual occurrence and background of the death. Furthermore, they argue that it is not conducive to debate what are true and false ideologies of martyrdom. In the context of this study though, it will become important to define martyrdom against necropolitics. This is to elucidate some understanding of the potential political risks that Abe's assassination may give rise to in terms of orchestrated assassinations.

 According to \citet{merriamweb}, it is "the suffering of death on account of adherence to a cause and especially to one's religious faith." Based on this, necropolitics will then be death that is not suffered, where the "victim" is willing to die. In this sense, Abe's death may be martyrdom, for he suffered death, where politicians interpreted  the assassination being an attack on democracy \citep{attackdemoc}. \citet{yilmaz} has analyzed in detail necropolitical icon heroes, finding that the idea of martyrdom is incredibly flexible and that the necropolitical use of martyrdom occurs mainly in non-democratic regimes for mass mobilization. Meanwhile, Abe's assassination has provided some insight into how necropolitics may occur in democracies.

While there are laws in place for when candidates pass away during elections, there have been no new laws formed for the passing of notable people with political influence who are not candidates during elections. This is likely due to there not being much precedence in events such as Abe’s assassination occurring. However, if one observes the results of the 2022 elections and how public sentiment was affected during such events, the future risks there being similar cases caused on purpose for political gain. 

Abe’s passing most certainly had a short-term impact on election sentiments, and with the event occurring extremely close to voting day, there is no doubt that the event was fresh on the minds of voters. As such, should a candidate be potentially interested in momentarily gaining the favor of the public’s opinion, they would theoretically organize an event in which a “martyr” is created immediately before voting day such that people are more swayed by pity and fear. The emotion of fear was observed to be especially affected for several hours following the assassination in tweets mentioning election-related terms.

The treatment of former Prime Minister Shinzo Abe by the Japanese government, following his assassination, greatly resembled that of a martyr. As noted by Masahiro Matsumura of St. Andrew’s University in an interview with CNBC, his death seemed to turn the election into a battle to “avenge” him, in the form of pity votes for the Liberal Democratic Party \citep{Matsumura}. Abe’s death does indeed seem to have influenced results, or at the very least, Twitter sentiments online. 

With regards to Abe's assassination, \citet{bob} may provide some foothold in explaining the election results and Twitter phenomenon observed. Among the five conclusions drawn on how state-sponsored removal of the head of movements will affect future mobilization, the fourth, where "movements whose goals are broader and whose leaders embody a shared
group identity are also more likely to inspire third-party support" offers interesting insight \citep{bob}. While Abe's death was not a state-sponsored removal, the Liberal Democratic Party (LDP) is a major political party in Japan whose leaders can be agreed to embody the shared identity of the LDP and their political goals. In this sense, Abe's death may have inspired third-party support for the candidate Abe was supporting and for the LDP. The sentiment changes seen on Twitter then could perhaps be said to be a reflection of the amount of "inspiration" Abe's death provided for the elections.

What is of great concern with this case study, is the orchestrated deaths of major political figures being exploited to manipulate election results to sway in favor of a certain party or candidate. As has been demonstrated by the election results following Abe’s death, the LDP does indeed seem to have gained ”pity votes,” leading them to secure 6 percent more seats following the election. Further supporting this claim, Japan is not the only nation to utilize martyrdom for political gain. As noted by Eli Alshech in 2008 in Die Welt des Islams, Hamas seemed to have gained success in the 2005 Municipal Elections in Palestine. As argued by them, the more martyrs there were, the more political foothold they gained within the Palestinian community \citep{Alsech}. Following Abe’s assassination, the Japanese Upper House in the National Diet saw the LDP grab 63 seats out of the 125 seats available \citep{japantimes}. Meanwhile, survey results following the elections showed that 15.1 percent of respondents had their votes swayed by the incident \citep{kyodo}. In future elections, in nations with voters who remain uncertain about who to cast their vote for, or nations with divided voters but low turnouts, Abe’s death may serve as a precedent to orchestrate necropolitics to encourage voters, especially those undecided, to cast revenge votes for the para-victim.

\section{Future Work}
The attention span of social media users is a promising avenue for further study. 
As has been discussed by \citet{holt}, “frequent social media use among young citizens can function as a leveler in terms of motivating political participation” \citep{holt}. Since the research of \citet{holt} focused on the 2010 Swedish national election campaign, it would be interesting to see if this holds true for young Japanese voters too.

As this study was conducted immediately after the assassination, the long-term effects of political martyrdom remain unclear. Future work could examine the lasting effects of Abe's death on Japanese politics and voter behavior.

In the future, it may be interesting to research how election candidates responded to and incorporated the assassination into their election campaigns. While it was outside the scope of this research, research into whether or not there were long-term effects will serve to answer some of the questions encountered throughout this research, especially those on necropolitics.

On the same note, the level of martyrdom of Abe, constructed by the government and Japanese society, is another avenue to research. As \citet{moskalenko} discussed, little is known about the relationship between the construction of martyrdom and its power to mobilize people. The case study of Abe may provide some foothold in gaining more clarity on this relationship. Given the initial results of this study, the amount of shock that political deaths provide in society may be a factor. To evaluate the validity of this factor, time series sentiment analysis for other political deaths would then become necessary.

Overall, this study opened more avenues for future research to gain a better understanding of political deaths and their online influence on society.
\subsection{Limitations}


Statistical research into verifying the results of works by researchers on the relationship between social media and election research could not be conducted. This is because there was very limited data on how users voted. While the utilization of voting preference polls conducted over phones was considered, this idea was discarded due to the user base of Twitter and those who are likely to answer phone polls being very different, in addition to difficulty in identification. Twitter has 58.95 million users in Japan \citep{statistaAI}, comprising 46.7 percent of the Japanese population as of January 2022. While no data on the distribution of Japanese Twitter users by age group was available, globally, 38.5 percent are 25-34, 20.7 percent are 35-49, 17.1 percent are 18-24, 17.1 percent are 50 or older, and 6.6 percent were 13-17 years of age \citep{statista2021}. On the other hand, in telephone polls, such as those conducted by NHK in July, 39.4 percent of those who answered were 70 or older, 16.3 percent were 60-69, 15 percent were 50-59, 10.1 percent were 40-49, 6.6 percent were 30-39, 5.2 percent were 20-29, while 1.9 percent were 18 or 19 years of age \citep{nhk}. With a large disparity in the distribution of Twitter user demographics and telephone poll answerers, it was not advisable to utilize phone polls in statistically determining the relationship between social media attitudes and elections. No other polling method was available in terms of societal data on voting preferences and attitudes toward elections in Japan.

While much care was taken to anonymize users and ensure that tweets used in the study were not leaked to 3rd parties, it must be noted that publicly available tweets can be easily searched and found. As such, some users may be easily identifiable. Similarly, deleted tweets have a retraceable digital footprint as observed by \citet{Almuhimedi}. For example, there are archival platforms, which will go unnamed in this paper, that allow one to easily retrieve deleted tweets. Additionally, certain tweets may also have been quoted by articles, further leaving them available despite being deleted by the user. 

\bibliographystyle{plainnat}
\bibliography{main}

\appendix\footnotesize
\label{sec:appendix}

\begin{figure*}
        \centering
        \includegraphics[width=\textwidth]{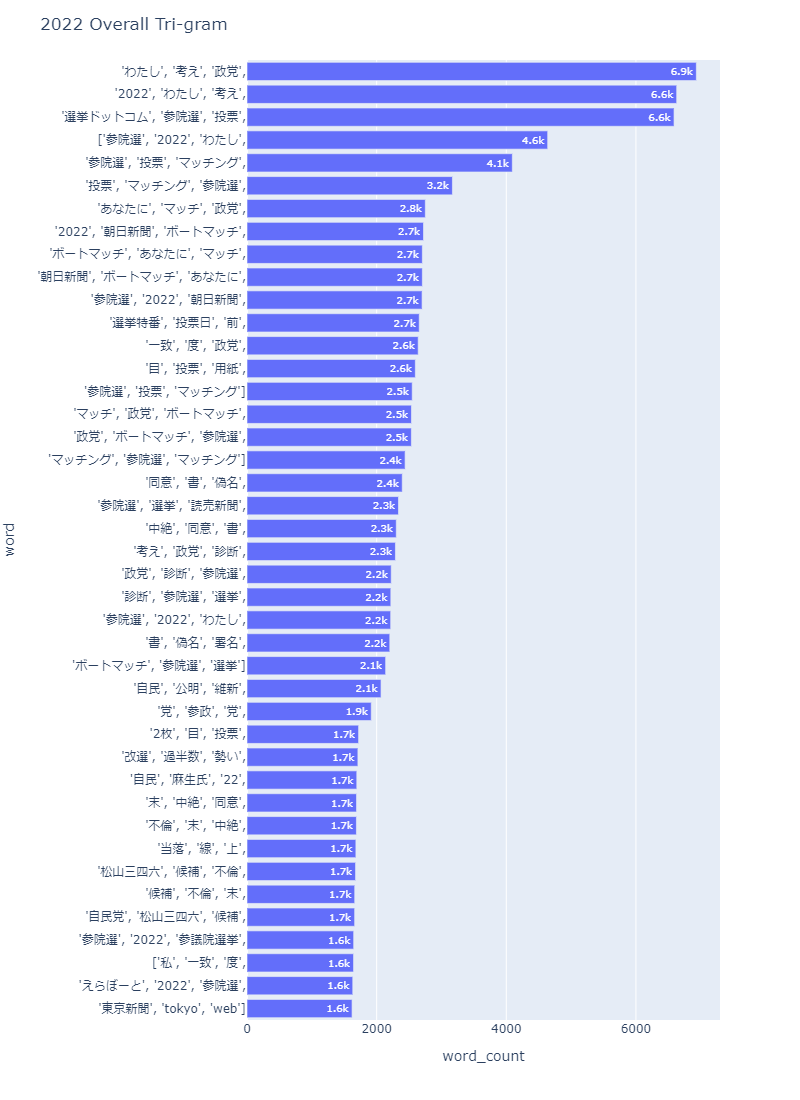}
        \caption{Trigram for Overall 2022 Election Period}
        \label{fig:2022 Overall Trigram}
\end{figure*}

\begin{figure*}
        \centering
        \includegraphics[width=\textwidth]{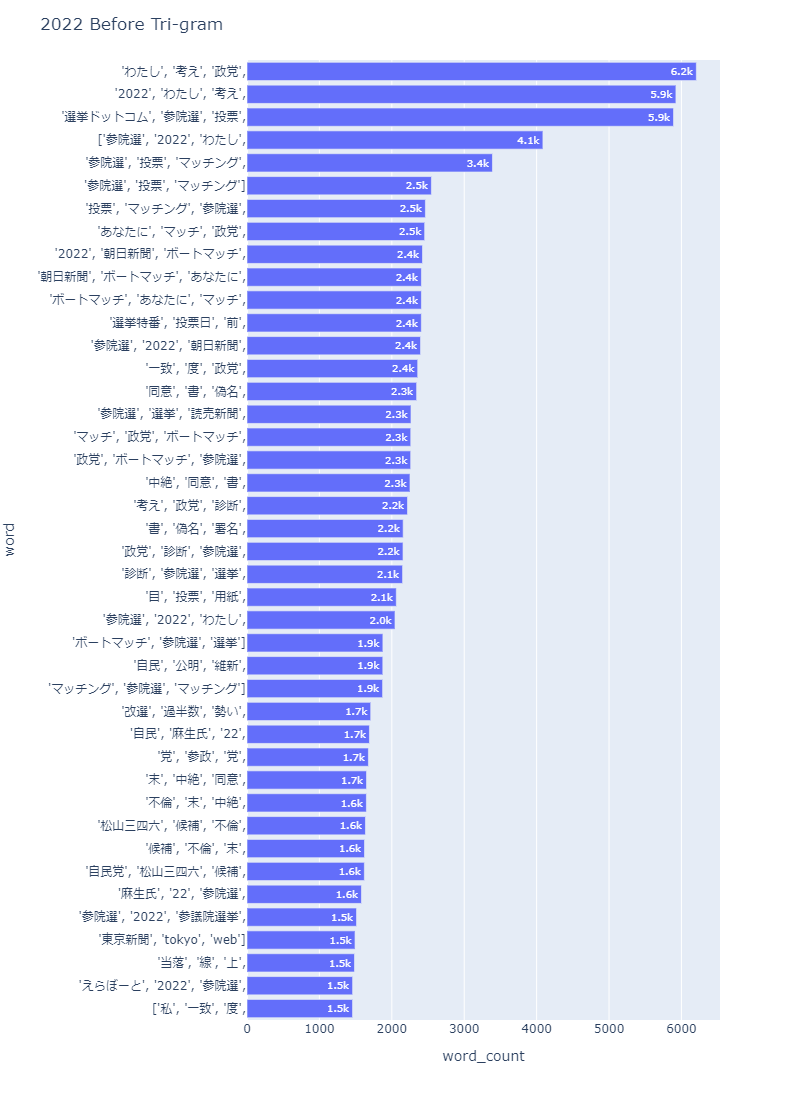}
        \caption{Trigram for Period Before Assassination}
        \label{fig:Before Assassination}
\end{figure*}

\begin{figure*}
        \centering
        \includegraphics[width=\textwidth]{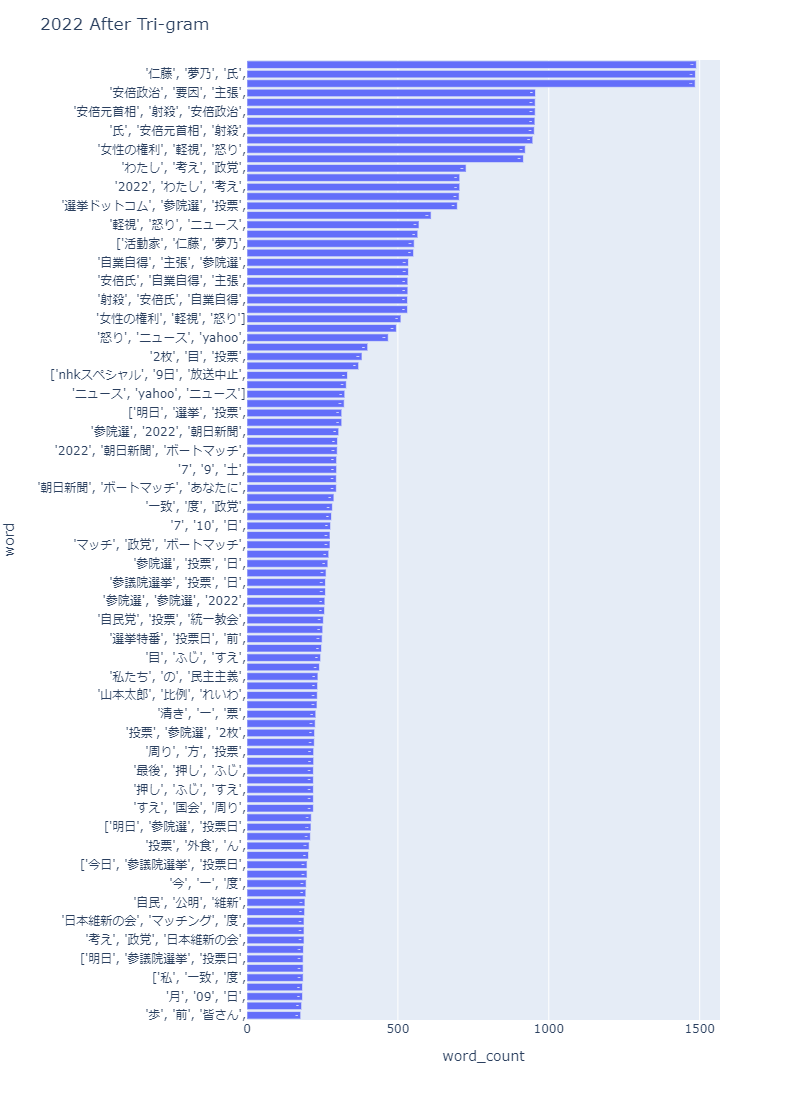}
        \caption{Trigram for Period After Assassination}
        \label{fig:After Assassination}
\end{figure*}



\end{document}